\documentclass[10pt, conference, compsocconf]{IEEEtran}
\usepackage{cite}

\ifCLASSINFOpdf
  \usepackage[pdftex]{graphicx}
\else
\fi
\usepackage[cmex10]{amsmath}
\usepackage{algorithmic}
\usepackage{url}

\usepackage[export]{adjustbox}
\usepackage{xcolor}

\usepackage{minibox}

\begin{document}

\title{Globally Optimal Object Tracking with Fully Convolutional Networks}

\author{Jinho Lee, Brian Kenji Iwana, Shouta Ide, Seiichi Uchida\\
Kyushu University, Fukuoka, Japan\\ Email: \{lee, brian, ide, uchida\}@human.ait.kyushu-u.ac.jp}

\maketitle

\begin{abstract}
Tracking is one of the most important but still difficult tasks in computer vision and pattern recognition. The main difficulties in the tracking field are appearance variation and occlusion. Most traditional tracking methods set the parameters or templates to track target objects in advance and should be modified accordingly. Thus, we propose a new and robust tracking method using a Fully Convolutional Network (FCN) to obtain an object probability map and Dynamic Programming (DP) to seek the globally optimal path through all frames of video. Our proposed method solves the object appearance variation problem with the use of a FCN and deals with occlusion by DP. We show that our method is effective in tracking various single objects through video frames.
\end{abstract}

\begin{IEEEkeywords}
Tracking, Fully Convolutional Network; Dynamic Programming

\end{IEEEkeywords}

\section{Introduction}

Object tracking is the subject of significant research in computer vision and pattern recognition. Research in this field contributes to state-of-the-art technologies such as service robots, automatic vehicles, security robots. However, to implement these technologies in society, there are still problems. In spite of many past researches, traditional object tracking methods still have limitations.

The primary difficulties of tracking are appearance changes (e.g. shooting direction, illumination, color, shape) and occlusion. Traditional methods track the target object with previously defined features. When the target object is changed, the features should be modified as well. This means that traditional tracking methods cannot track various target objects through a fixed algorithm.

To tackle appearance changes, we present the use of an artificial neural network.  
The concept of neural networks is to use interconnected nodes to work together to obtain information.
The first advent of neural networks was 1970s and it consists of a number of layers of perceptrons\cite{1}.
In contrast to the past when neural networks faced computational requirement problems, recent work have achieved state-of-the-art successes. 
This is attributed to the introduction of GPUs, increased data availability, and processing power.

A Convolutional Neural Network (CNN) \cite{2} is a type of artificial neural network which apply learned convolutional kernels to layer inputs to transform input data as a method of feature extraction. The general idea is to use the learned features for image recognition. CNNs are powerful visual models that yield hierarchies of features. CNNs are not only improving for whole image classification field, but also making progress on local tasks with structured output. These include advances in object detection and local correspondence.

Furthermore, Fully Convolutional Networks (FCN) \cite{3} are composed entirely of convolutional layers and has been used recently in simultaneous segmentation and classification. An FCN is trained end-to-end, pixels-to-pixels. It takes input of arbitrary size and produce a correspondingly-sized output with efficient inference and learning. Both inference and learning are performed on the whole image by feedforward computation and backpropagation. The pixelwise prediction and learning can be obtained by upsampling the subsampled pooling layers.

To deal with an occlusion, we apply Dynamic Programming (DP) \cite{4,5} to obtain the globally optimal tracking path through all frames. DP is one of the most fundamental optimization techniques and has been used for object tracking. It is a non-greedy algorithm to estimate the global optimal path over matching items in a sequence. Because of this nature, DP-based tracking is robust to occlusion, which degrades the tracking performance of greedy algorithms.

The contribution of this paper is to demonstrate that our proposed tracking method, which combines a FCN and DP, can deal with the difficulties of the tracking task. We obtain probability maps of the target object which correspond to each input frame through the use of a FCN. DP, then, follows the highest values of the probability map efficiently under a slope constraint to obtain the smooth tracking path.

The remaining of this paper is organized as follows. Section~\ref{related} provides related works in detection and tracking methods with artificial neural networks. Section~\ref{fcn} and \ref{dp} elaborates on FCNs and DP and also the details of the proposed method. In Section~\ref{results}, we confirm the proposed method and analyzed the experimental results. Finally, Section~\ref{conclusion} draws the conclusion.

\section{Related Works}
\label{related}

The purpose of the object tracking is locating a target object accurately through a connected sequence of frames.
It has a rich history with many methods proposed by past researchers. Despite the many proposed methods, it is not yet a solved task. 

The most simple method of object tracking is to use a sliding window and obtaining a probability map for all possible sliding window sizes over the frames. However, this method is inefficient because of the time-cost. Region-based Convolutional Neural Networks (R-CNN) \cite{6} improve this by proposing regions of interest, but is still slow because it also has to compute each forward pass for each object proposal, without sharing computation. To solve that problem, SPPnet \cite{7} was proposed but has notable drawbacks. The main drawback is that the fine-tuning algorithm cannot update the convolutional layers which precede the spatial pyramid pooling. After advent of SPPnet, Fast R-CNN \cite{8} and Faster R-CNN \cite{9,10} proposed  applying a ROI pooling layer and RPN (Region Proposal Network) as improves upon this.

There, also, are various related tracking research using various neural networks. For example, neural networks have been used for human tracking \cite{11}, recognition of hand gestures \cite{12}, pedestrian detection \cite{13}, robust tracking with a single modified CNN \cite{14}, multitarget tracking \cite{15,16}, and a tracking method combining CNNs and optical flow \cite{17}. Deconvolutional networks and a FCNs have been used for segmentation \cite{18,19}. 3D convolutional neural networks \cite{20} have been applied to object tracking as well.

\section{Fully Convolutional Networks}
\label{fcn}

\begin{figure}[t]
	   		\begin{center}

				\includegraphics[width=8cm]{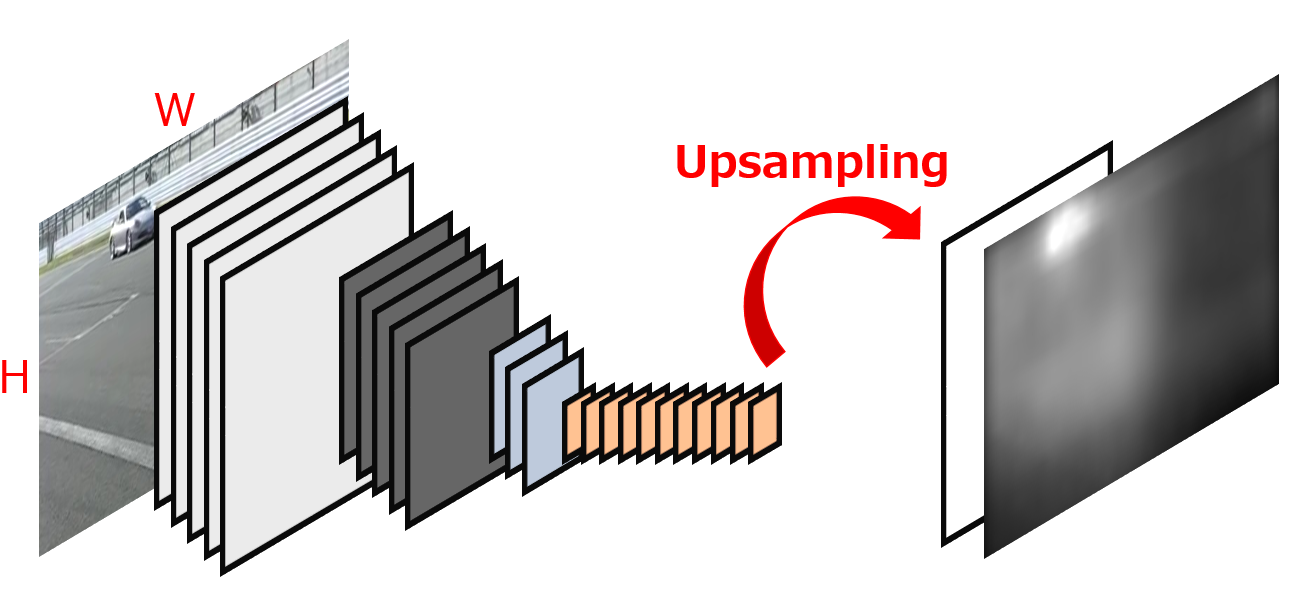}
				\vspace{-2mm}					


				\caption {A diagram of the structure of a FCN. The initial input is downsampled through pooling and upsampled back to the original image size. Pixelwise prediction is done on the result.}
				\label{zu1}

			\end{center}
\end{figure}

Before the introduction of FCNs, a CNN was used in various fields. Normal CNNs are made up of three components: convolutional layers, pooling layers, and fully-connected layers. However, a FCN views the fully-connected layers as convolutional layers and uses upsampling layers with the skip connections to learn multiscaled features. Fig. \ref{zu1} illustrates the structure of a FCN and how upsampling can produce same-sized inputs and outputs.

In the convolutional layers, feature maps are produced by applying a number of the filters from inputs to the layer. The filters are learned and weighted through the repeated backpropagation. The resulting feature maps are fed into an pooling layers to reduce the computational time for future layers. After the downsampling is done by the pooling layers, they are fed into an activation function. The output of the layer is next passed to further convolutional layers and pooling layers until they are reduced to ${1\times1}$ convolutions. 

\begin{figure}[t]
	   		\begin{center}

				\includegraphics[width=9cm]{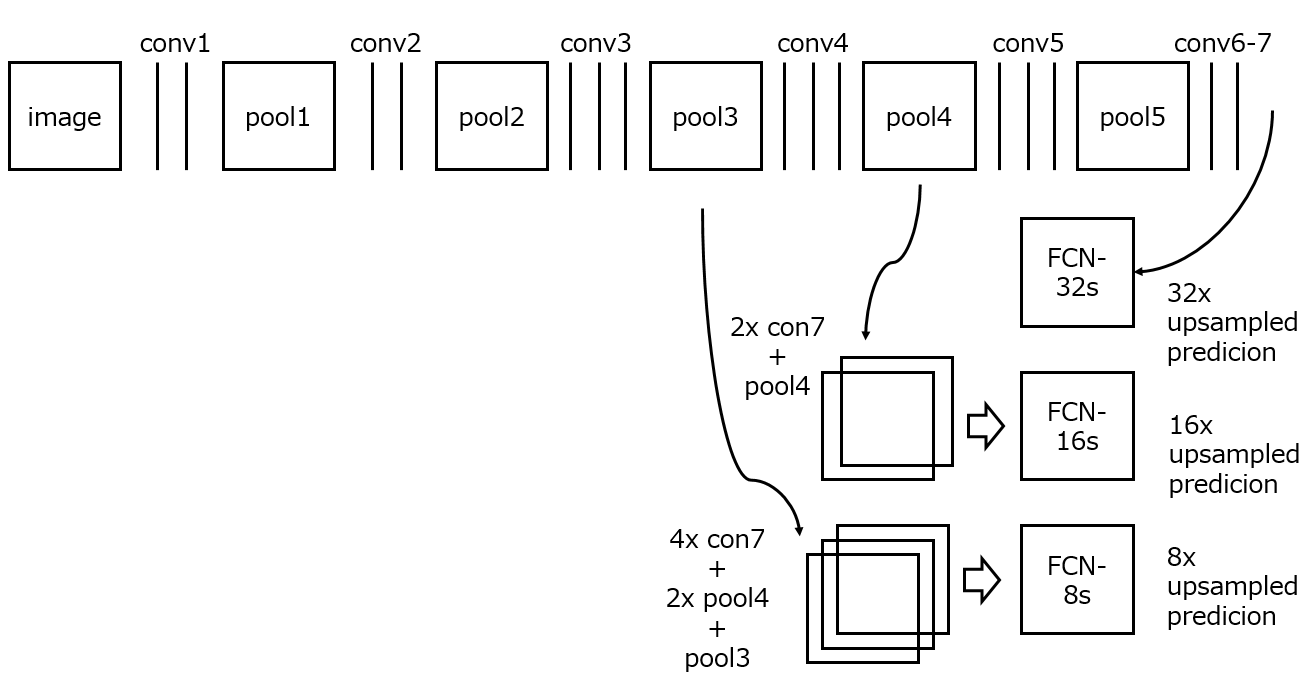}
				\vspace{-2mm}					


				\caption {Skip connections combine information from the the lower and higher layers. FCN-32s: upsamples stride 32 predictions back to pixels in a single step, FCN-16s: combines predictions from both the final layer and the pool4 later, at stride 16, FCN-8s: additional predictions from pool3 layer.}
				\label{zu2}

			\end{center}
\end{figure}

Next, the outputs, namely the feature maps, are upsampled through layers until the result is the same size as the original input image where pixelwise classification is performed. However, by itself, this process produces unsatisfactory results. To solve this, links between the low-level fine layers and the high-level coarse layers are constructed. These so called \textit{skip connections} combine information from the fine layers and course layers. The structure of the skip connections are shown in Fig. \ref{zu2}. We use the output of FCN-8s.

Compared to sliding window based region networks, FCNs are much faster because FCNs only require a single feedforward for computation. The other merit is that compared to methods using a fixed sliding window, it can deal with the scale changes of target object. 

FCNs can create probability maps for quick segmentation and classification which also deals with appearance variation, such as scale changes of the target object, shape, color, shooting direction, and illumination.

\section{Dynamic Programming}
\label{dp}

In our research, we use DP to obtain the most optimal path through all frames. It is one of the basic and essential optimization techniques and has been used for object tracking. 

\begin{figure}[t]
	   		\begin{center}

				\includegraphics[width=9cm]{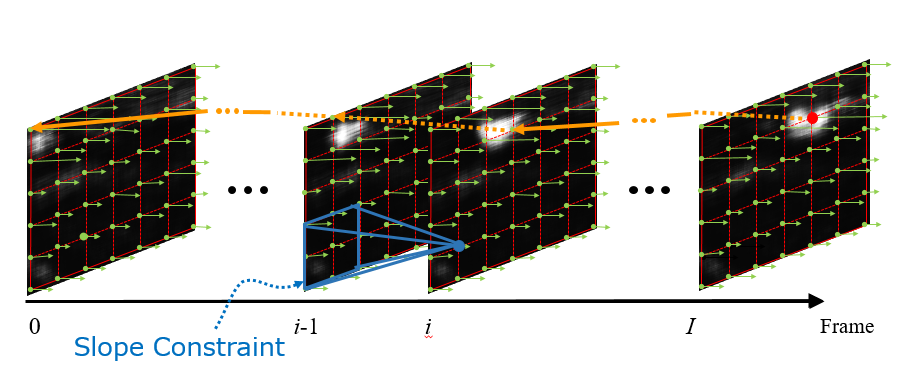}
				\vspace{-2mm}					


				\caption {DP Tracking structure- Green arrows: values of the probability of a target object, Blue rectangle: slope constraint, Red plot: the highest sum value in the last probability map, Orange: back-tracking path (the most globally optimal tracking path)}
				\label{zu3}

			\end{center}
\end{figure}

To apply DP to our method, we start by creating probability maps of each of the frames. For each pixel on the probability map, we find the highest value from the previous frame given a slope constraint. We create a cumulative DP map from the search. This process is continued by iterating across all of the frames. On the final frame, we select the highest value from the cumulative DP map and backtrack along the previous frames selecting the route with the highest values. An illustration of this process is shown in Fig. \ref{zu3}. The resulting path is the estimated optimal tracking path. Tracking an object through the local maximum limited by the slope constraint is what provides DP with the robustness with occlusion.

\section{Experimental results}
\label{results}

\subsection{Dataset preparation}
\begin{figure}[t]
	   		\begin{center}

				\includegraphics[width=8cm]{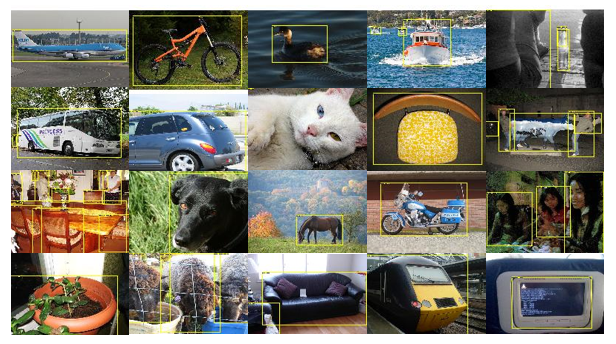}
				\vspace{-2mm}					


				\caption {20 categories examples of the VOC2012 dataset to train a FCN}
				\label{zu0}

			\end{center}
\end{figure}

We used the VOC2012 \cite{21} dataset and trained a FCN on 20 categories. The training dataset had 11,530 images containing 27,450 ROI annotated objects and 6,929 segmentations. The categories are as follows, {\it person, bird, cat, cow, dog, horse, sheep, airplane, bicycle, boat, bus, car, motorbike, train, bottle, chair, dining table, potted plant, sofa, and tv/monitor}. In Fig. \ref{zu0}, we show examples of VOC2012 dataset.

\subsection{Evaluation}

To demonstrate that our proposed method can track the target object in high accuracy, we evaluated our proposed method with following the rule of the Visual Tracker Benchmark \cite{22}. However, there are two conditions to evaluate.

\begin{enumerate}
 \item We evaluate sequences that have a target object in one of the 20 categories of VOC2012. Our proposed tracking method can only detect objects in the 20 categories.   
 \item We evaluate our proposed method by comparing the precision of established methods. The precision is defined as the average Euclidean distance between the center locations of the tracked targets and the manually labeled ground truths. The average center location error over all of the frames of one sequence is used to summarize the overall performance for that sequence. To show the performance efficiently, we calculate the percentage of frames whose estimated location is within the given threshold distance of the ground truth.
\end{enumerate} 

As we mentioned in condition 1), our proposed method can detect only trained objects through 20 categories of VOC2012. According to that condition, we select the 12 sequences for experiments: {\it CarScale, Coke, Couple, Crossing, David3, Jogging1\&2, MotorRolling, MountainBike, Walking1\&2, Woman}. In the 12 sequences, there are attributes which include the difficulties of tracking e.g. illumination variation, scale variation, occlusion, fast motion, rotation, and low resolution. Although we limit the target sequences, our proposed method does not need any template for tracking, in contrast that traditional trackers need to change the template according to change of appearance.

Several sequences contain multiple targets although our current DP tracking only allows to track a single object. When we try to track with using sequences containing multiple targets, our proposed method does not know which object to track because of the nature of DP. To overcome this, we can initialize the method by identifying the target object location on the first probability map. This initialization not just allows us to handle multi-object scenarios, but also increases the accuracy of single-object tracking. To compare accuracy between with initialization and non-initialization, we separate two categories (i.e. multiple sequences: {\it CarScale, Coke, Couple, Crossing, David3, Jogging1\&2, MotorRolling, MountainBike, Woman}, single sequences: {\it CarScale, Coke, Couple, Crossing, David3, MotorRolling, MountainBike, Walking1\&2, Woman}) and do the experiments respectively.

\subsection{Comparison Experiments \& Analysis}
Tracking methods can be divided into offline like our tracking method and online tracking. However, we compare our proposed method to online tracking methods in order to show the performance, because there is no comparable offline trackers which is released.\footnote{In the future research, we will apply the network flow \cite{23} which is online tracking method to FCNs and present performance of combination of FCNs and network flow.} We compare to the top 5 traditional trackers introduced in the Visual Tracker Benchmark (i.e. Structured Output Tracking with Kernels (Struck) \cite{24}, a sparsity-based tracker (SCM) \cite{25}, P-N Learning tracker (TLD) \cite{26}, Context tracker (CXT) \cite{27}, and Visual Tracking Decomposition (VTD) \cite{28}). 

\begin{figure}[t]
	   		\begin{center}

				\includegraphics[width=8cm]{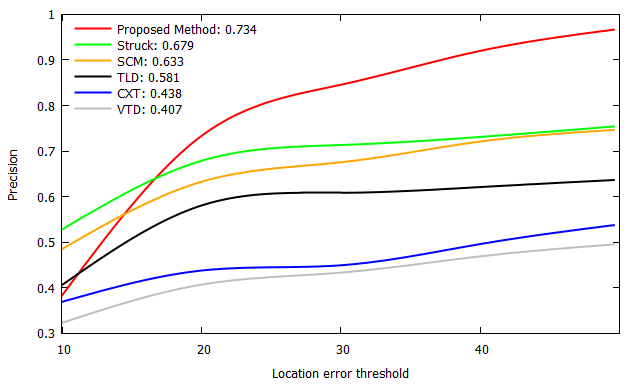}
				\vspace{-2mm}					


				\caption {Summarize the all results with initialization by using multiple sequences: {\it CarScale, Coke, Couple, Crossing, David3, Jogging1\&2, MotorRolling, MountainBike, Walking1\&2, Woman}. The score listed in the legend means the precision score at a 20 pixel threshold. In Fig.\ref{zu6}, we can check that our proposed method possesses the higher performance for all thresholds above 20.}
				\label{zu6}

			\end{center}
\end{figure}

{\it Proposed Method} is the evaluation of the proposed method using one-pass evaluation (OPE) introduced in the Visual Tracker Benchmark. It is a common way to evaluate trackers which runs them throughout a test sequence with initialization from the ground truth in the first frame and calculate the average precision error. We summarize all of the precision results with initialization with the percentage of frames whose estimated location is within the given threshold distance of the ground truth in Fig. \ref{zu6}. Proposed Method is the results when identifying the target object on the first frame. The proposed method has a higher performance for all threshold above 20. This means that proposed method can track target object with little mistakes, however it does not outperform in aspect of fine accuracy. The reason why the proposed method can track with little mistakes, is that sets the slope constraint to prohibit tracked location from moving rapidly. When the target object is enough big, our proposed method might track the point which is far from the center point of target object, because FCNs do not always obtain the highest probability value in which is close to center point of target object. That is why precision of our proposed method is lower that that of the other trackers for thresholds below 20. 

\begin{figure}[t]
	   		\begin{center}

				\includegraphics[width=9cm]{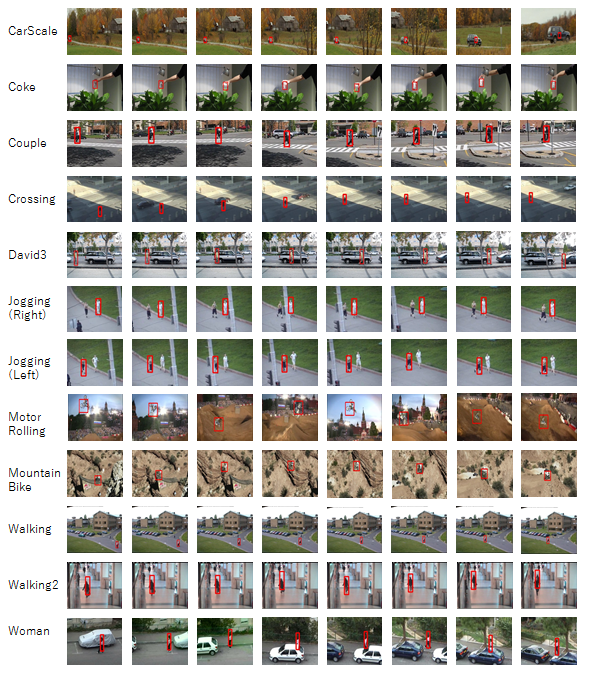}
				\vspace{-2mm}					


				\caption {Sequences which are evaluated by Proposed Method with fixed parameters  following the OPE. The red rectangles are centered around the tracking result for visualization purposes.}
				\label{zu4}

			\end{center}
\end{figure}


There are three advantages of our proposed method. First, the proposed method does not need any template for tracking. For traditional tracking, it is necessary to prepare a template to track a target object when the object changes. However, for our proposed method, it is unnecessary under the condition that the class of the target object is trained by the FCN. Shown in Fig. \ref{zu4}, our proposed method has a high accuracy in tracking various objects. However, objects not trained by the FCN (i.e. not included in VOC2012) cannot be tracked. 

\begin{figure}[t]
	   		\begin{center}

				\includegraphics[width=8cm]{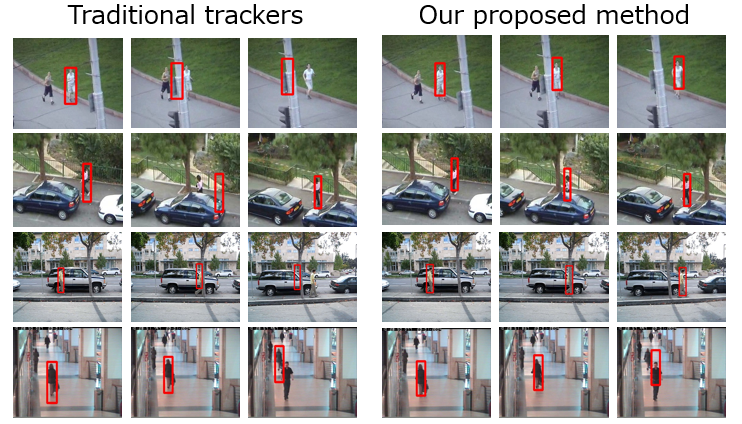}
				\vspace{-2mm}					


				\caption {Examples of the proposed method dealing with appearance variation and occlusion. Examples (left) are results of traditional trackers (i.e. Struck tracker results \cite{24}: 1st, 3rd, 4th sequences, CXT tracker results \cite{27}: 2nd sequence). Examples (right) is the results of our proposed methods.}
				\label{zu5}

			\end{center}
\end{figure}

Second, the proposed method does not need to modify the parameters even if appearance of the target object is changed. Our proposed method can deal with appearance variation by learning numerous features of the target object. As shown Fig. \ref{zu5}, although there are different people as the target object, such as variation of shooting direction, color, scale and motion, in each sequence, our proposed method  correctly tracks each person without any modification of parameters.

Third, the proposed method can deal with occlusion because DP seeks the global optimal  path over all of the frames. Shown in Fig. \ref{zu5}, our proposed method can track the target object be occluded, in contrast to Struck, the top method in the Visual Tracker Benchmark. However, when the objects of same category are occluded, our proposed method does not know which object should track like the final sequence in Fig. \ref{zu5}.

\begin{figure}[t]
	   		\begin{center}

				\includegraphics[width=8cm]{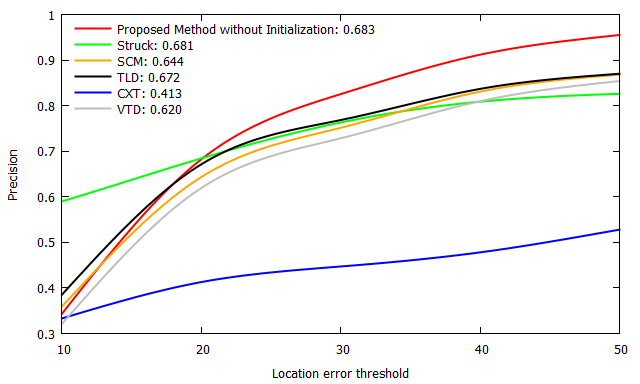}
				\vspace{-2mm}					


				\caption {Compare the results of the Proposed Method without Initialization and those of traditional trackers with initialization by using single sequences: {\it CarScale, Coke, Couple, Crossing, David3, MotorRolling, MountainBike, Walking1\&2, Woman}. The score listed in the legend means the precision score at a 20 pixel threshold. Through comparing to traditional trackers, our proposed method can track more correctly even if it is not given initialization.}
				\label{zu7}

			\end{center}
\end{figure}

Also, we did experiments to compare qualitative evaluations between with initialization (i.e. using multiple sequences) and non-initialization (i.e. using single sequences).  Also In Fig. \ref{zu7}, we show comparing the results of our proposed method without initialization and those of traditional trackers with initialization by using single sequences. \textit{Proposed Method without Initialization} is the results without the same initialization. As seen in Fig. \ref{zu7}, the proposed method is accurate  even if it runs without initialization. It also outperforms the traditional trackers above threshold 20.

In the tracking field, it is also important to compare the time requirement for computation. Table 1 shows the average frames per second (fps) of each tracker in OPE running on a PC with Intel i7 6850CPU (3.6GHz) and GTX-1080$\times$2. Offline which is involved in our tracking method, is robust to occlusion, because it searches for the tracking path looking after over all frames. However, it has a drawback that requires a numerous time cost in contrast to online tracking. So this is why our tracking method is slower than the other methods. It should be noted that the time cost of the proposed method is sensitive to size of input and slope constraint.

\begin{table}[t]
 \begin{center}
   \caption{Frame rate (fps) comparing between our proposed method and traditional trackers.}
    \begin{tabular}{|c|c|} \hline
     Tracker & Frame rate (fps)  \\ \hline
     Proposed Method & 0.211  \\ \hline 
     Struck & 25.1 \\ \hline
     SCM & 0.771 \\ \hline
     TLD & 36.2 \\ \hline
     CXT & 23.6 \\ \hline
     VTD & 9.82 \\ \hline
    \end{tabular}
  \end{center}
\end{table}

\section{Conclusion}
\label{conclusion}

In this paper, we presented that the combination of FCNs and DP to track the objects. We showed that it is possible to track an object in situations that have occlusion and appearance variation as a highlight of our research. Using the proposed method also has a high accuracy compared to the traditional tracking methods without modifying parameters and template. These are a big contribution in tracking field. However, when the FCN cannot detect target objects or when objects with similar classes occlude the target, our proposed tracking method might track incorrectly.

Future research will be that make the FCN more accurately detect objects by augmentation of the dataset and using the FCN-4s which is upsampled output from the sum of upsampled FCN-8s and pool2 layer to get the more high resolution probability map. Second, by applying Flownet that makes a FCN learn information of optical flow, we can increase the tracking accuracy even if similar object appears. Finally, we will automatically make the tracked bounding box in each frame and compare with the other competition results. Evaluation by overlapped bounding area between ground truth and tracked bounding box can measure the tracking performance more efficient.


\let\oldbibliography\thebibliography
\renewcommand{\thebibliography}[1]{\oldbibliography{#1}
\setlength{\itemsep}{2mm}}

\end{document}